%% file: samplepaper.tex
\documentclass[runningheads]{llncs}
\usepackage{graphicx}
\usepackage{epsfig} 
\usepackage{mathptmx} 
\usepackage{times} 
\usepackage{amsmath} 
\usepackage{amssymb}  

\usepackage[utf8]{inputenc} 
\usepackage[T1]{fontenc}    
\usepackage{hyperref}       
\usepackage{url}            
\usepackage{booktabs}       
\usepackage{amsfonts}       
\usepackage{nicefrac}       
\usepackage{microtype}      

\usepackage{graphicx}
\usepackage{subcaption}
\usepackage{stmaryrd} 
\usepackage{breakcites}
\usepackage{authblk}
\usepackage{multicol}
\usepackage{comment}

\newcommand{\tim}[1]{\textcolor{black}{#1}}

\usepackage[colorinlistoftodos]{todonotes}

\begin{document}
\title{\LARGE \bf
Marginal Replay vs Conditional Replay for Continual Learning
}
\titlerunning{Marginal Replay vs Conditional Replay for CL}
%

\author{Timoth\'ee Lesort\inst{1, 2} \orcidID{0000-0002-8669-0764} \and
Alexander Gepperth\inst{3} \orcidID{0000-0003-2216-7808} \and
Andrei Stoian\inst{2} \orcidID{0000-0002-3479-9565} \and
David Filliat\inst{1} \orcidID{0000-0002-5739-1618} }

\authorrunning{T. Lesort et al.}
%
\institute{Flowers Laboratory (ENSTA ParisTech \& INRIA), France \and
 Thales, Theresis Laboratory, France \and
 Fulda University of Applied Sciences, Germany}

\maketitle              
\begin{abstract}
  We present a new replay-based method of continual classification learning that we term "conditional replay" which generates samples and labels together by sampling from a distribution conditioned on the class. We compare conditional replay to another 
  replay-based continual learning paradigm (which we term "marginal replay") that generates samples independently of their class and assigns labels in a separate step. 
  The main improvement in conditional replay is that labels for generated samples need not be inferred, which reduces the margin for error in complex continual classification learning tasks. We demonstrate the effectiveness of this approach using novel and standard benchmarks constructed from MNIST and FashionMNIST data, and compare to the regularization-based \textit{elastic weight consolidation} (EWC) method \cite{Kirkpatrick2016, shin2017continual}.

\end{abstract}

\section{Introduction}
\begin{figure}
    \centering
    \includegraphics*[viewport=0in 0in 4.8in 1.3in,width=0.59\textwidth]{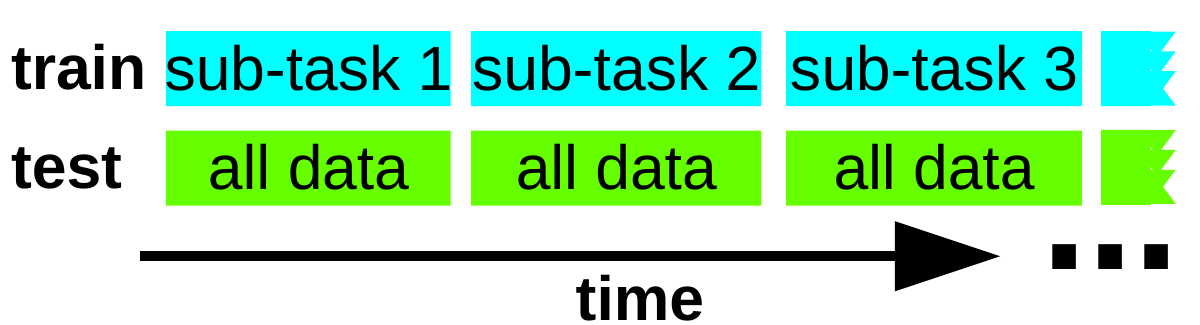}
    \includegraphics[width=0.40\textwidth]{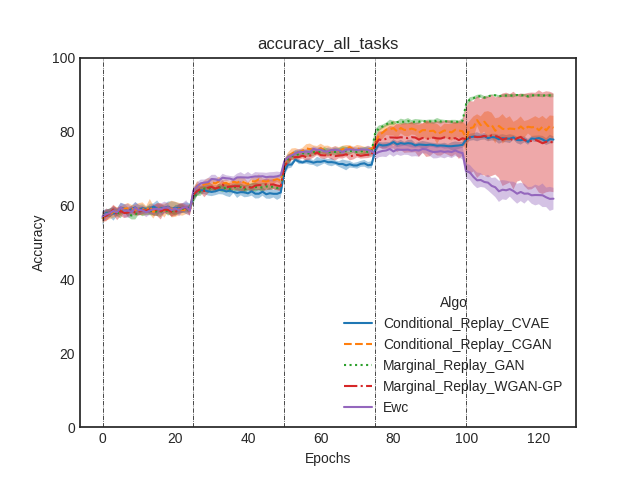}
    \caption{Left: The problem setting of continual learning as investigated in this article. DNN models are trained one after the other on a sequence of sub-tasks (of which three are shown here), and are continuously evaluated on a test set consisting of the union of all sub-task test sets. This gives rise to results as shown exemplarily on the right-hand side of the figure, i.e., plots of test set accuracy over time for different models, where boundaries between sub-tasks (5 in this case) are indicated by vertical lines.
    \label{fig:my_label}
    }
\end{figure}
This contribution is in the context of incremental, continual or lifelong learning, subject that is gaining increasing recent attention \cite{parisi2018continual,gepperth2016incremental} and for which a variety of different solutions have recently been proposed (see below). Briefly put, the problem consists of repeatedly re-training a deep neural network (DNN) model with new sub-tasks, or continual learning tasks (CLTs), (for example: new visual classes) over long time periods, while avoiding the abrupt degradation of previously learned abilities that is known under the term "catastrophic interference" or "catastrophic forgetting"  \cite{gepperthICANN,french,gepperth2016incremental}. Please see Fig.~\ref{fig:my_label} for a visualization of the problem setting. Is has long been known that catastrophic forgetting (CF) is a problem for connectionist models  \cite{french} of which modern DNNs are a specialized instance, but only recently there have been efforts to propose workable solutions to this problem for deep learning models  \cite{lee2017overcoming,Kirkpatrick2016,selfless,DBLP:journals/corr/abs-1805-10784,3862}. A recent article \cite{pfuelb2019a} demonstrates empirically that most proposals fail to eliminate CF when common-sense application constraints are imposed (e.g., restricting prior access to data from new sub-tasks, or imposing constant, low memory and execution time requirements). 

One aspect of the problem seems to be that gradient-based DNN training is greedy, i.e., it tries to optimize all weights in the network to solve the current task only. Previous tasks, which are not represented in the current training data, will naturally be disregarded in this process. While approaches such as  \cite{Kirkpatrick2016,lee2017overcoming} aim at "protecting" weights that were important for previous tasks, one can approach the problem from the other end and simply include samples from previous tasks in the training process each time a new task is introduced. 

This is the \textit{generative replay} approach, which is in principle model-agnostic, as it can be performed with a variety of machine learning models such as decision trees, support vector machines (SVMs) or deep neural networks (DNNs). 
It is however unfeasible for, e.g., embodied agents or embedded devices performing object recognition, to store all samples from all previous sub-tasks. Because of this, 
generative replay proposes to train an additional machine learning model (the so-called \textit{generator}). Thus, the "essence" of previous tasks comes in the form of trained generator parameters which usually require far less space than the samples themselves. 
A downside of this and similar approaches is that the time complexity of adapting to a new task is not constant but depends on the number of preceding tasks that shouldbe replayed. Or, conversely, if continual learning should be  performed at constant time complexity, only a fixed amount of samples can be generated, and thus there will be forgetting, although it won't be catastrophic.

\tim{In this paper we decide to investigate two different types of generative models : Generative adversarial networks (GAN) and variational auto-encoder (VAE). On one hand GAN are known to generate samples of high quality but on the other hand VAE directly maximize the likelihood of the learned distribution while training. It was then interesting to experiment both of them to compare their performances.}

This article proposes and evaluates a particular method for performing replay using DNNs, termed "conditional replay", which is similar in spirit to  \cite{shin2017continual} but presents important conceptual improvements (see next section). The main advantage of conditional replay is that samples can be generated conditionally, i.e., based on a provided label. Thus, labels for generated samples need not be inferred in a separate step as other replay-based approaches, e.g., \cite{shin2017continual}, which we term \textit{marginal replay} approaches. Since inferring the label of a generated sample inevitably requires the application of a possibly less-than-perfect classifier, avoiding this step conceivably reduces the margin for error in complex continual learning tasks. 
\tim{The paper organization is the following, first we introduce the paper contributions and the related works, secondly we describe the methods used as well as the benchmarks, thirdly we present the paper experiments, fourthly we show and discuss our results and in a last section we conclude the paper.}

\subsection{Contribution}\label{sec:contr}
The original contributions of this article can be summarized as follows: 
\begin{itemize}
    \item \textbf{Conditional replay as a method for continual classification learning} We experimentally establish the advantages of conditional replay in the field of continual learning by comparing conditional and marginal replay models on a common set of benchmarks. 
    \item \textbf{Improvement of marginal learning} We furthermore propose an improvement of marginal replay as proposed in  \cite{shin2017continual} by using generative adversarial networks (GANs, see \cite{goodfellow2014generative}). 
    \item {New experimental benchmarks for generative replay strategies} To measure the merit of these proposals, we use two experimental settings that have not been previously considered for benchmarking generative replay: rotations and permutations. In addition, we promote the ''10-class-disjoint'' task as an important benchmark for continual learning as it is impossible to solve for purely discriminative methods (at no time, samples from different classes are provided for training so no discrimination can happen). 
    \item \textbf{Comparison of generative replay to EWC} We show the principled advantage that generative replay techniques have with respect to regularization methods like EWC in a "one class per task" setting, which is after all a very common setting in practice and in which discriminatively trained models strongly tend to assign the same class label to every sample regardless of content.
\end{itemize}

\input{Tables/hyperparameters.tex}
\begin{figure*}
    \centering
    \begin{subfigure}[b]{0.15\textwidth}
        \includegraphics[width=\textwidth]{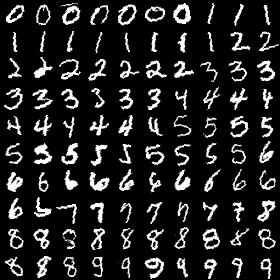}
        \caption{sub-task 0}
        \label{fig:mnist_rotations_0}
    \end{subfigure}
    \begin{subfigure}[b]{0.15\textwidth}
        \includegraphics[width=\textwidth]{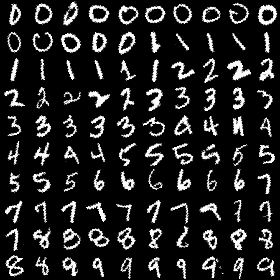}
        \caption{sub-task 1}
        \label{fig:mnist_rotations_1}
    \end{subfigure}
        \begin{subfigure}[b]{0.15\textwidth}
        \includegraphics[width=\textwidth]{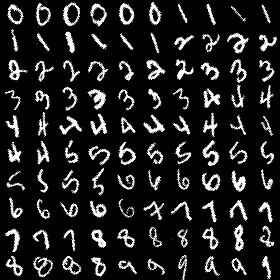}
        \caption{sub-task 2}
        \label{fig:mnist_rotations_2}
    \end{subfigure}
    \begin{subfigure}[b]{0.15\textwidth}
        \includegraphics[width=\textwidth]{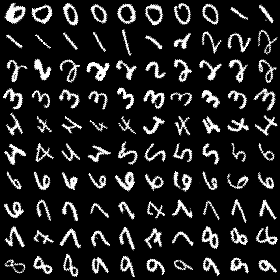}
        \caption{sub-task 3}
        \label{fig:mnist_rotations_3}
    \end{subfigure}
    \begin{subfigure}[b]{0.15\textwidth}
        \includegraphics[width=\textwidth]{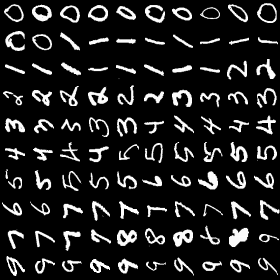}
        \caption{sub-task 4}
        \label{fig:mnist_rotations_4}
    \end{subfigure}
  \caption{MNIST training data for rotation sub-tasks.}
\end{figure*} 

\begin{figure*}
    \centering
    \begin{subfigure}[b]{0.9\textwidth}
        \includegraphics[width=\textwidth]{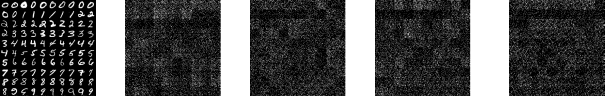}
        \label{fig:mnist_permutation_train}
    \end{subfigure}
   
  \caption{MNIST training data for permutation-type CLTs.}
\end{figure*} 
\subsection{Related work}
The field of continual learning is growing and has been recently reviewed in, e.g.,  \cite{parisi2018continual,gepperth2016incremental}. In the context of neural networks, principal recent approaches include ensemble methods \cite{ren2017life,fernando2017pathnet,mallya2018packnet,rusu2016progressive,yoon2017lifelong,aljundi2017expertGate,serra2018overcoming,li2018learning}, regularization approaches \cite{Kirkpatrick2016,lee2017overcoming,selfless,DBLP:journals/corr/abs-1805-10784,Srivastava2013,Hinton2012,aljundi2018memory,liu2018rotate,chaudhry2018riemannian,gepperth2019matrix}, dual-memory systems  \cite{kemker2017fearnet,rebuffi2017icarl,gepperth2015bio}, distillation-based approaches \cite{shmelkov2017incremental,li2018learning,kim2018keep} and generative replay methods \cite{shin2017continual,kemker2017fearnet,lesort2018generative,kamra2017deep,wu2018memory}. In the context of single-memory DNN methods, regularization approaches are predominant: whereas it was proposed in  \cite{Goodfellow2013} that the popular Dropout regularization can alleviate catastrophic forgetting, the EWC method  \cite{Kirkpatrick2016} proposes to add a term to the DNN energy function that protects weights that are deemed to be important for the previous sub-task(s). Whether a weight is important or not is determined by approximating and analyzing the Fisher information matrix of the DNN.
A somewhat related approach is pursued with the incremental moment matching (IMM, see  \cite{lee2017overcoming}) technique, where weights are transferred between DNNs trained on successive sub-tasks by regularization techniques, and the Fisher information matrix is used to "merge" weights for current and past sub-tasks.
Other regularization-oriented approaches are proposed in  \cite{selfless,Srivastava2013} which focus on enforcing sparsity of neural activities by lateral interactions within a layer, or in  \cite{DBLP:journals/corr/abs-1805-10784}. Concerning recent advances in generative replay improving upon  \cite{shin2017continual}:
Several works propose the use of generative models in continual learning of classification tasks \cite{Kamra17, wu18incremental, wu2018memory, Shah18} but their results does not provide comparison between different types of generative models.
 \cite{2018arXiv181209111L} propose a conditional replay mechanism similar to the one investigated here, but their goal is the sequential learning of data generation and not classification tasks.
Generally, each approach to continual learning has its advantages and disadvantages:
\begin{itemize}
\item ensemble methods suffer from little to no interference between present and past knowledge as usually different networks or sub-networks are allocated to different learning tasks. The problem with this approach is that, on the one hand, model complexity is not constant, and more seriously, that the 
task from which a sample is coming from must be known at inference time in order to select the appropriate (sub-)network.
\item regularization approaches are very diverse: in general, their advantage is simplicity and (often) a constant-time/memory behavior w.r.t. the number of tasks. However, the impact of the regularizer on continual learning performance is difficult to understand, and several parameters need to be tuned whose significance is unclear (i.e., the strengths of the regularization terms)
\item distillation approaches can achieve very good robustness and continual learning performance, but either require the retention of past samples, or the occurrence of samples from past classes in current training data to be consistent. Also, the strength of the various distillation loss regularizers needs to be tuned, usually by cross-validation.
\item generative replay and dual-memory systems show very good and robust continual learning performance, although time complexity of learning depends on the number of previous tasks for current generative replay methods. In addition, the storage of weights for a sufficiently powerful generator may prove very memory-consuming, so this approach cannot be used in all settings.
\end{itemize}
\section{Methods}
A basic notion in this article is that of a continual (or sequential) learning task (CLT or SLT, although we will use the abbreviation CLT in this article), denoting a classification problem that is composed of two or more sub-tasks which are presented sequentially to the model in question. Here, the CLTs are constructed from two standard visual classification benchmarks: MNIST and Fashion MNIST, either by dividing available classes into several sub-tasks, or by performing per-sample image processing operations that are identical within, and different between, sub-tasks. All continual learning models are then trained and evaluated in an identical fashion on all CLTs, and performances are compared by a simple visual inspection of classification accuracy plots.
\subsection{Benchmarks}
\par\noindent
\textbf{MNIST}~  \cite{LeCun1998} is a common benchmark for computer vision systems and classification problems. It consists of gray scale 28x28 images of handwritten digits (ten balanced classes representing the digits 0-9). The train, test and validation sets contain 55.000, 10.000 and 5.000 samples, respectively. \\

\par\noindent
\textbf{Fashion MNIST}~  \cite{Xiao2017} consists of grayscale 28x28 images of clothes.   We choose this dataset because it claims to be a ``more challenging classification task than the simple MNIST digits data  \cite{Xiao2017}'' while having the same data dimensions, number of classes, balancing properties and number of samples in train, test and validation sets. 

\subsection{Continual learning tasks (CLTs)}
All CLTs are constructed from the underlying MNIST and FashionMNIST benchmarks, 
so the number of samples in train and test sets for each sub-task depend on the precise way of constructing them, see below.
\par 
\begin{figure*}
    \centering
    \begin{subfigure}[b]{0.12\textwidth}
        \includegraphics[width=\textwidth]{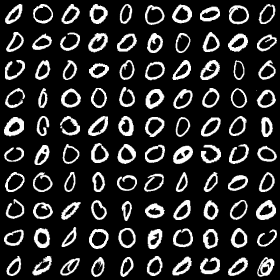}
        \caption{Sub-task 0 }
        \label{fig:mnist_disjoint_0}
    \end{subfigure}
    \begin{subfigure}[b]{0.12\textwidth}
        \includegraphics[width=\textwidth]{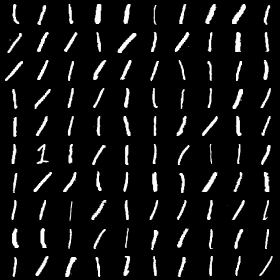}
        \caption{sub-task 1 }
        \label{fig:mnist_disjoint_1}
    \end{subfigure}
        \begin{subfigure}[b]{0.12\textwidth}
        \includegraphics[width=\textwidth]{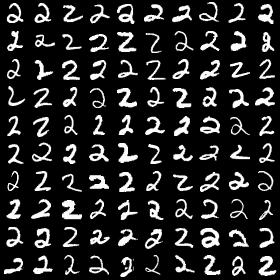}
        \caption{sub-task 2 }
        \label{fig:mnist_disjoint_2}
    \end{subfigure}
    \begin{subfigure}[b]{0.12\textwidth}
        \includegraphics[width=\textwidth]{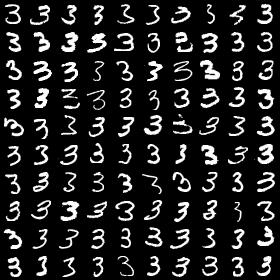}
        \caption{sub-task 3 }
        \label{fig:mnist_disjoint_3}
    \end{subfigure}
    \begin{subfigure}[b]{0.12\textwidth}
        \includegraphics[width=\textwidth]{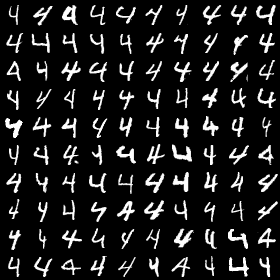}
        \caption{sub-task 4 }
        \label{fig:mnist_disjoint_4}
    \end{subfigure}
    
        \begin{subfigure}[b]{0.12\textwidth}
        \includegraphics[width=\textwidth]{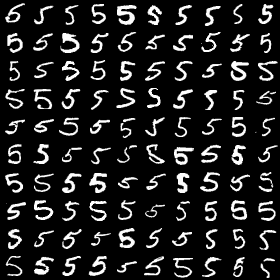}
        \caption{sub-task 5 }
        \label{fig:mnist_disjoint_5}
    \end{subfigure}
    \begin{subfigure}[b]{0.12\textwidth}
        \includegraphics[width=\textwidth]{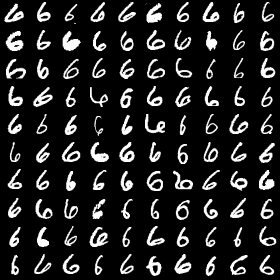}
        \caption{sub-task 6 }
        \label{fig:mnist_disjoint_6}
    \end{subfigure}
        \begin{subfigure}[b]{0.12\textwidth}
        \includegraphics[width=\textwidth]{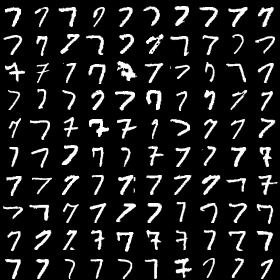}
        \caption{sub-task 7 }
        \label{fig:mnist_disjoint_7}
    \end{subfigure}
    \begin{subfigure}[b]{0.12\textwidth}
        \includegraphics[width=\textwidth]{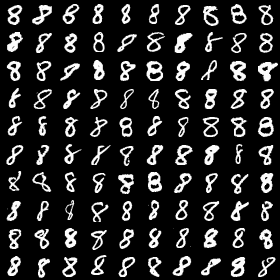}
        \caption{sub-task 8}
        \label{fig:mnist_disjoint_8}
    \end{subfigure}
    \begin{subfigure}[b]{0.12\textwidth}
        \includegraphics[width=\textwidth]{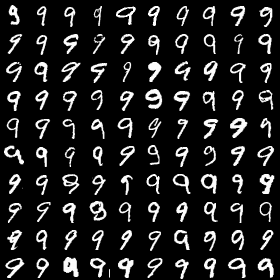}
        \caption{sub-task 9 }
        \label{fig:mnist_disjoint_9}
    \end{subfigure}
   
   \caption{Samples of MNIST training data for the disjoint CLT. Each sub-task adds one more visual class, a principle which carries over identically to FashionMNIST.}
\end{figure*}
\par\noindent
\textbf{Rotations}~New sub-tasks are generated by choosing a random rotation angle $\beta \in [0,\pi/2]$ and then performing a 2D in-plane rotation on all samples of the original benchmark. As both benchmarks we use contain samples of 28x28 pixels, no information loss is introduced by this procedure. We limit rotation angles to $\pi/2$ because larger rotations could mix MNIST classes like 6 and 9. Each sub-task in rotation-based CLTs contains all 10 classes of the underlying benchmark, leading to 55.000 and 10.000 samples, respectively, in the train and test sets of each sub-task.\\

\par\noindent
\textbf{Permutations}~New sub-tasks are generated by defining a random pixel permutation scheme, and then applying it to each data sample of the original benchmark. Each sub-task in permutation-based CLTs contains all 10 classes of the underlying benchmark, leading to 55.000 and 10.000 samples, respectively, in the train and test sets of each sub-task.\\

\par\noindent
\textbf{Disjoint classes}~For each benchmark, this CLT has as many sub-tasks as there are classes in the benchmark (10 in this article). Each sub-task contains the samples of a single class, i.e., roughly 6.000 samples in the train set and 1.000 samples in the test set. As the classes are balanced for both benchmarks, this does not unduly favor certain classes. This CLT presents a substantial challenge for machine learning methods since a normal DNN would, for each sub-task, learn to map all samples to a single class label irrespective of content. Selective discrimination between any two classes is hard to obtain except if replay is involved, because then a classifier actually "sees" samples from different classes at the same time.
\subsection{Models}
In this article, we compare a considerable number of deep learning models: unless otherwise stated, we 
employ the Rectified Linear Unit (ReLU) transfer function, cross-entropy loss for classifier training, and the Adam optimizer.

\textbf{EWC} We re-implemented the algorithm described in  \cite{Kirkpatrick2016}, choosing
two hidden layers with 200 neurons each.\\

\par\noindent
\textbf{Marginal replay}~ {In the context of classification, the \textit{marginal replay}  \cite{2018arXiv181209111L, shin2017continual, wu2018memory} method works as follows : 
For each sub-task $t$, there is a dataset $D_t$, a classifier $C_t$, a generator $G_t$ and a memory of past samples composed of a generator $G_{t-1}$ and a classifier $C_{t-1}$. The latter two allow the generation of artificial samples $D_{t-1}$ from previous sub-tasks.
Then, by training $C_t$ and $G_t$ on $D_t$ and $D_{t-1}$, the model can learn the new sub-task $t$ without forgetting old ones.}
At the end of the sub-task, $C_t$ and $G_t$ are frozen and replace $C_{t-1}$ and $G_{t-1}$.
In the default setting, we use the generator for marginal replay in a way that ensures a balanced distribution of classes from past sub-tasks $D_{t-1}$, see also Fig.~\ref{fig:distribution}. This is achieved by  choosing a predetermined number of samples $N$ to be added for all sub-tasks t, and letting the generator produce $tN$ previous samples at sub-task $t$. Thus, 
the number of generated samples increases linearly over time. We choose to evaluate two different models for the generator: WGAN-GP as used in  \cite{shin2017continual} and the original GAN model  \cite{NIPS2014_5423} since it is a competitive baseline \cite{lesort2018training}.\\

\par\noindent
\textbf{Conditional replay}~ The conditional replay method is derived from \textit{marginal replay}: instead of saving a classifier and a generator, the algorithm only saves a generator that can generate conditionally (for a certain class).
Hence, for each sub-task $t$, there is a dataset $D_t$, a classifier $C_t$ and two generators $G_t$ and $G_{t-1}$.
The goal of $G_{t-1}$ is to generate data from all the previous sub-tasks during training on the new sub-task. Since data is generated conditionally, samples automatically have a label and do not require a frozen classifier. We follow the same strategy as for marginal replay (previous paragraph) for choosing the number of generated samples at each sub-task. However, conditional replay does not require this: it can, in principle, keep the number of generated samples constant for each sub-task since it is trivially possible to generate a balanced distribution of $\frac{N}{t}$ samples per class, from $t$
different classes, via conditional sample generation.
$C_t$ and $G_t$ learn from generated data $D_{t-1}$ and $D_t$. At the end of a sub-task $t$, $C_t$ is able to classify data from the current and previous sub-tasks, and $G_t$ is able to sample from them also. We choose to use two different popular conditional models : CGAN described in  \cite{mirza2014conditional} and CVAE \cite{NIPS2015_5775}.
\begin{figure*}[h!]
    \centering
    \begin{subfigure}[b]{0.45\textwidth}
        \includegraphics[width=\textwidth]{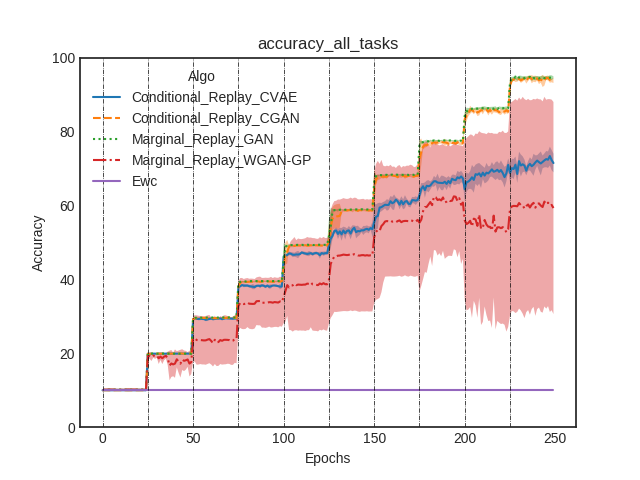}
        \caption{accuracy for MNIST disjoint CLT}
        \label{fig:mnist_disjoint_all_task_accuracy}
   \end{subfigure}
    \begin{subfigure}[b]{0.45\textwidth}
        \includegraphics[width=\textwidth]{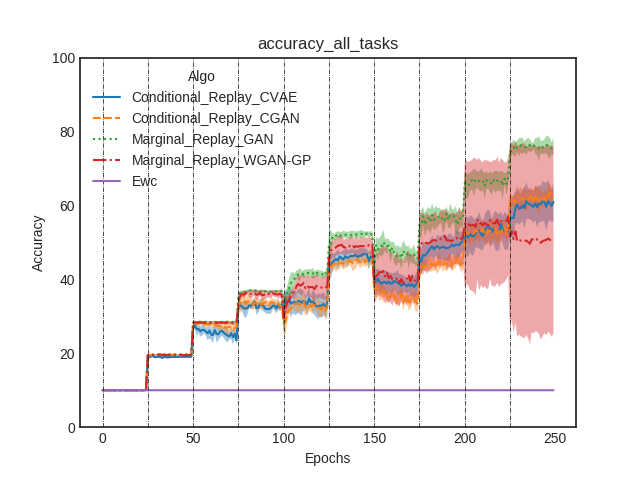}
        \caption{accuracy for Fashion MNIST disjoint CLT}
        \label{fig:fashion_disjoint_all_task_accuracy}
   \end{subfigure}
   
       \centering
    \begin{subfigure}[b]{0.45\textwidth}
        \includegraphics[width=\textwidth]{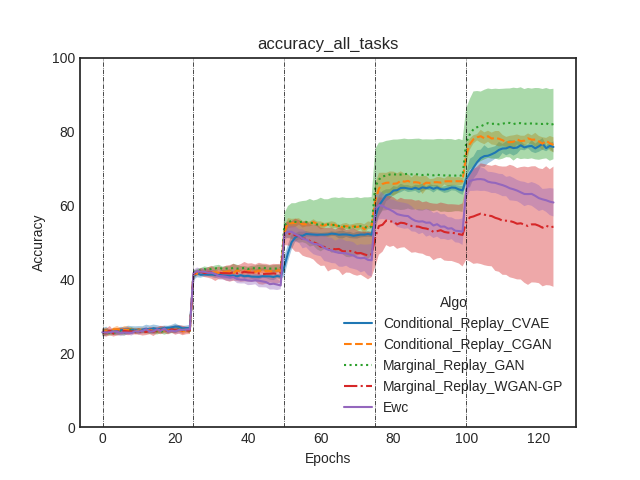}
        \caption{accuracy for MNIST permutation CLT}
        \label{fig:mnist_permutations_all_task_accuracy}
    \end{subfigure}
   \begin{subfigure}[b]{0.45\textwidth}
       \includegraphics[width=\textwidth]{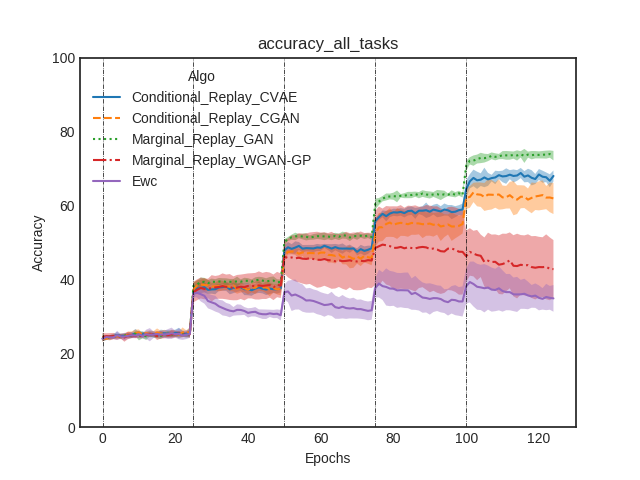}
        \caption{accuracy for Fashion MNIST permutation CLT}
        \label{fig:fashion_permutations_all_task_accuracy}
   \end{subfigure}

       \centering
    \begin{subfigure}[b]{0.45\textwidth}
        \includegraphics[width=\textwidth]{Figures/mnist_rotations_all_task_accuracy}
        \caption{accuracy for MNIST rotation CLT}
        \label{fig:mnist_rotations_all_task_accuracy}
    \end{subfigure}
   \begin{subfigure}[b]{0.45\textwidth}
       \includegraphics[width=\textwidth]{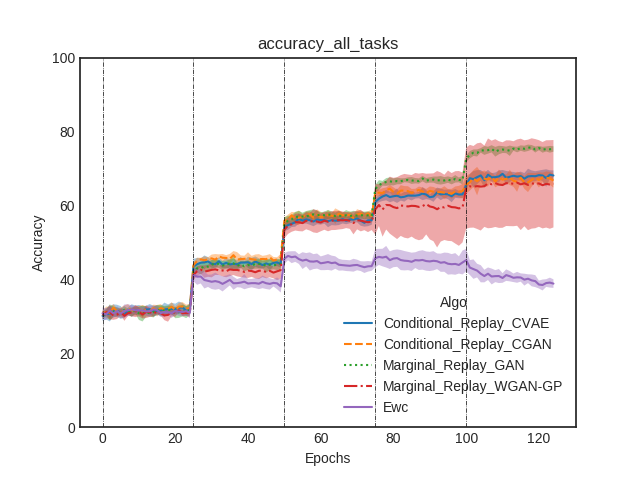}
        \caption{accuracy for Fashion MNIST rotation CLT}
        \label{fig:fashion_rotations_all_task_accuracy}
   \end{subfigure}

   \caption{Test set accuracies during training on different CLTs, shown for all sub-tasks (indicated by dotted lines).}
    \label{fig:all_task_accuracy}
\end{figure*} 
\begin{figure*}[h]
    \centering
    \begin{subfigure}[b]{0.45\textwidth}
       \includegraphics[width=\textwidth]{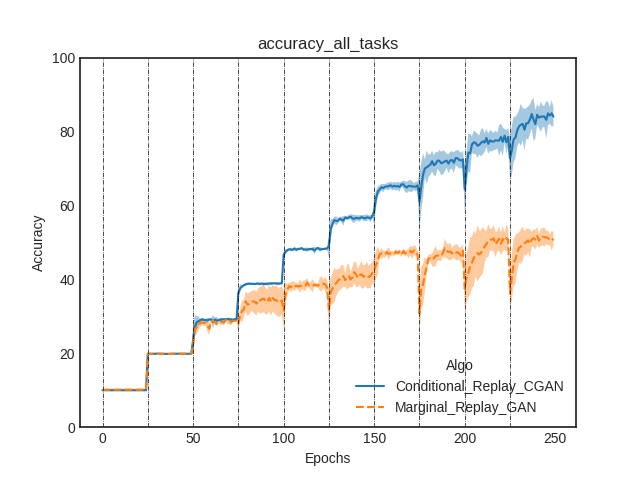}
        \caption{Unbalanced MNIST Disjoint}
        \label{fig:unbalanced_mnist}
    \end{subfigure}
   \begin{subfigure}[b]{0.45\textwidth}
       \includegraphics[width=\textwidth]{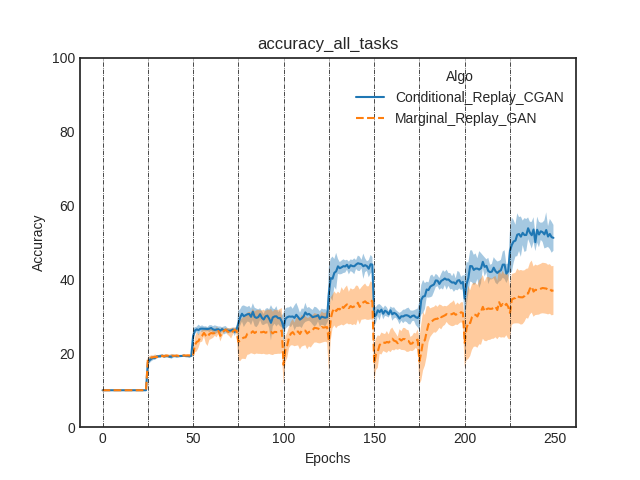}
        \caption{Unbalanced Fashion Disjoint}
        \label{fig:unbalanced_fashion}
   \end{subfigure}
   
       \centering
    \begin{subfigure}[b]{0.45\textwidth}
       \includegraphics[width=\textwidth]{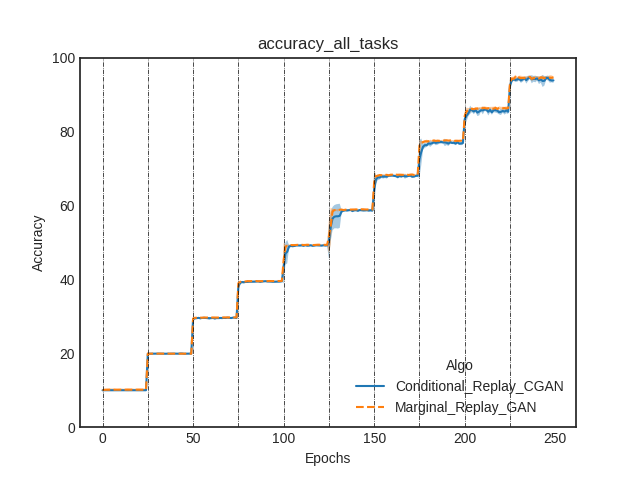}
        \caption{Balanced MNIST Disjoint}
        \label{fig:balanced_mnist}
    \end{subfigure}
   \begin{subfigure}[b]{0.45\textwidth}
       \includegraphics[width=\textwidth]{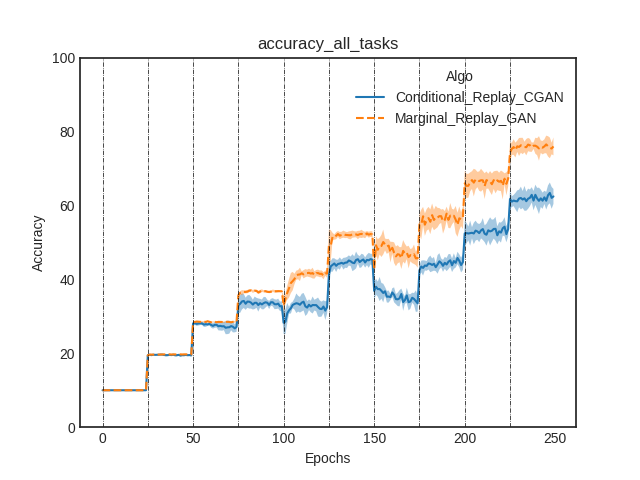}
        \caption{Balanced Fashion Disjoint}
        \label{fig:balanced_fashion}
   \end{subfigure}
   
   \caption{\label{fig:bal}
   We compare final accuracy when the ratio between size of old task and size of new task is 1 (balanced) or 1/10 (unbalanced, factor was chosen empirically).}
\end{figure*} 
\section{Experiments}
We conduct experiments using all models and CLTs described in the previous section. Each class (regardless of the CLT) is presented for 25 epochs, 
Results are presented either based on the time-varying classification accuracy on the \textit{whole} test set, or on the class (from the test set) that was presented first. In the first case, accuracy should ideally increase over time and reach its 
maximum after the last class has been presented. In the second case, accuracy \tim{should be stable if the model does not forgot} or decrease over time, reflecting that some information about the first class is forgotten. 
We distinguish two major experimental goals or questions: 
\begin{itemize}
    \item Establishing the performance of the newly proposed methods (marginal replay with GAN, conditional replay with CGAN or CVAE) w.r.t. the state of the art. To this effect, we conduct experiments that increase the number of generated samples over time in a way that ensures an effectively balanced class distribution (see Fig.~\ref{fig:distribution}). We do this both for marginal and conditional replay in order to ensure a fair comparison, although technically conditional replay can generate balanced distribution even with a constant number of generated samples.
    \item demonstrating the advantages of conditional w.r.t. marginal replay strategies, especially when only few samples can be generated, thus obtaining a skewed class distribution for marginal replay (see Fig.~\ref{fig:distribution}). 
\end{itemize} 
Results shedding light on the first question are presented in Fig.~\ref{fig:all_task_accuracy} (showing classification accuracy on whole test set over time, see Fig.~\ref{fig:first} for accuracy on first sub-task), whereas the second question is addressed in Fig.~\ref{fig:bal} for the disjoint CTL only due to space limitations. 
\section{Results and discussion}
From the experiments described in the previous section, we can state the following principal findings:\\

\begin{figure}
    \centering
    \includegraphics[width=0.44\textwidth]{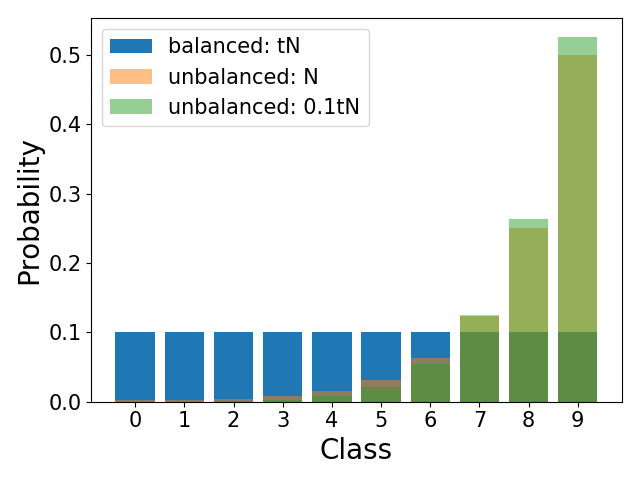}
    \caption{Why marginal replay must linearly increase the number of generated samples: distribution of classes produced by the generator of a marginal replay strategy after sequential training of 10 sub-tasks (of 1 class each). This essentially corresponds to the "disjoint" type of CLTs. Shown are three cases: "\textit{balanced}: $tN$" (blue bars) where $tN$ samples are generated for each sub-task $t$, "unbalanced: $N$" (orange bars) where the number of generated samples is constant and equal to the number of newly trained samples $N$ for each sub-task, and "unbalanced: $0.1 tN$" where $0.1tN$ samples are generated. We observe that, in order to ensure a balanced distribution of classes, the number of generated samples must be re-scaled, or, in other words, must increase linearly with the number of sub-tasks.
    }
    \label{fig:distribution}
\end{figure}

\par\noindent
\textbf{Replay methods outperform EWC} As can be observed from Fig. \ref{fig:all_task_accuracy}, the 
novel methods we propose (marginal replay with GAN and WGAN-GP, conditional replay with CGAN and conditional replay with CVAE) outperform EWC, on all CLTs, sometimes by a large margin.
Particular attention should be given to the performance of EWC: while generally acceptable for rotation and permutation CLTs, it completely fails for the disjoint CLT. This is due to the fact that there is only one class in each sub-task, making EWC try to map all samples to the currently presented class label regardless of input, since no replay is available to include samples from previous sub-tasks (as outlined before in Sec.~\ref{sec:contr}).\\

\par
\begin{figure*}[t!]
    \centering
    \begin{subfigure}{0.31\textwidth}
        \includegraphics*[width=\textwidth]{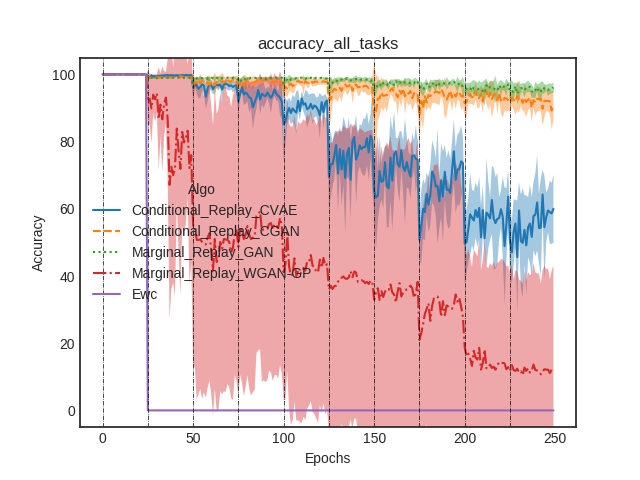}
        \caption{MNIST: disjoint CLT}
        \label{fig:mnist_disjoint_first_task_accuracy}
    \end{subfigure}
    \begin{subfigure}{0.31\textwidth}
        \includegraphics*[width=\textwidth]{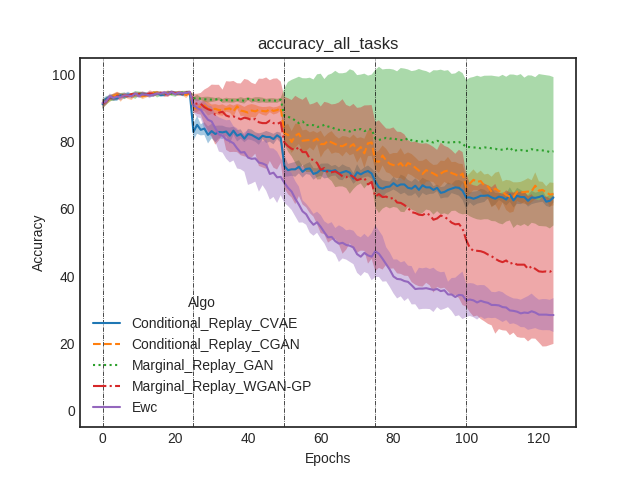}
        \caption{MNIST: permutation CLT}
        \label{fig:mnist_permutations_first_task_accuracy}
    \end{subfigure}
    \begin{subfigure}{0.31\textwidth}
        \includegraphics*[width=\textwidth]{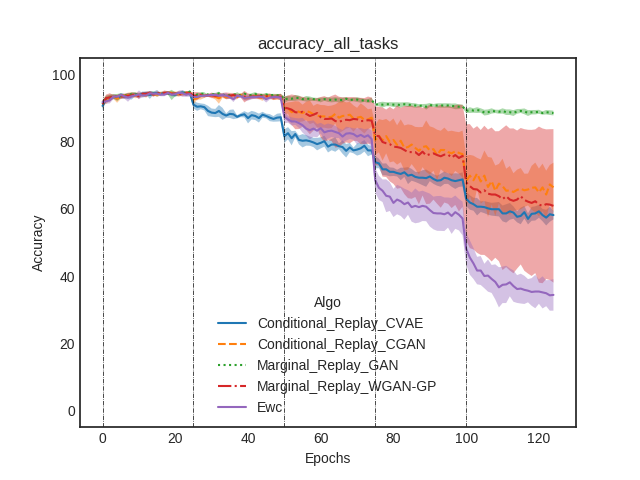}
        \caption{MNIST: rotation CLT}
        \label{fig:mnist_rotations_first_task_accuracy}
    \end{subfigure}
    \centering
    
   \begin{subfigure}[b]{0.32\textwidth}
       \includegraphics[width=\textwidth]{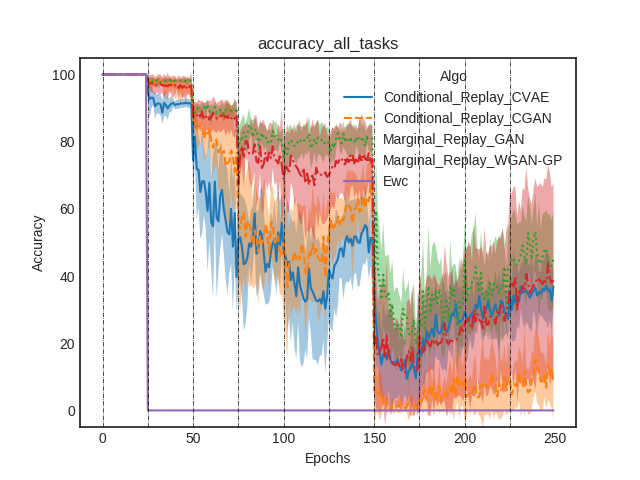}
        \caption{Fashion MNIST: disjoint CLT}
        \label{fig:fashion_disjoint_first_task_accuracy}
   \end{subfigure}
   \begin{subfigure}[b]{0.32\textwidth}
       \includegraphics[width=\textwidth]{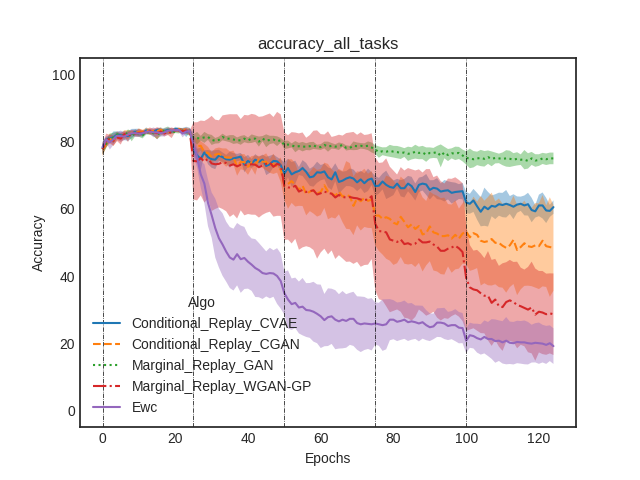}
        \caption{Fashion MNIST: permutation CLT}
        \label{fig:fashion_permutations_first_task_accuracy}
   \end{subfigure}
   \begin{subfigure}[b]{0.32\textwidth}
       \includegraphics[width=\textwidth]{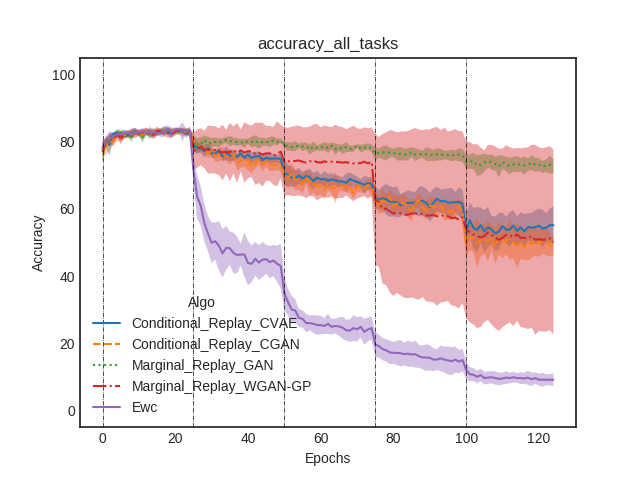}
        \caption{Fashion MNIST: rotation CLT}
        \label{fig:fashion_rotations_first_task_accuracy}
   \end{subfigure}
   \caption{\label{fig:first}
   Comparison of the accuracy of each approach on the first sub-task. This is another, very intuitive measure of how much is forgotten during continual learning. Means and standard deviations computed over 8 seeds.}
\end{figure*}

\noindent
\textbf{Marginal replay with GAN outperforms WGAN-GP}
The clear advantage of GAN over WGAN-GP is the higher stability of the generative models.
This is not only observable in Fig. \ref{fig:all_task_accuracy}, but also when measuring performance on the first sub-task only during the course of 
continual learning (see Fig.\ref{fig:first}).\\

\par\noindent
\textbf{Conditional replay can be run at constant time complexity}
A very important point in favour of conditional replay is run-time complexity, as expressed by the number of samples that need to be generated each time a new sub-task is trained. Since the generators in marginal replay strategies generate samples regardless of class, the distribution of classes will be proportional to the distribution of classes during the last training of the generator, which leads to an unbalanced class distribution over time, with the oldest classes being strongly under-represented (see Fig. \ref{fig:distribution}). This is
avoided by increasing the number of generated samples over time for marginal replay, leading to a balanced class distribution (see also Fig. \ref{fig:distribution}) while vastly increasing the number of samples. Conditional replay, on the other hand, 
can selectively generate samples from a defined class, thus constructing a class-balanced dataset without needing to increase the number of generated samples over time. In the interest of accuracy, it can of course make sense to
increase the number of generated samples over time, just as for marginal replay. This, however, is a deliberate choice and not something required by conditional replay itself.\\

\par\noindent
\textbf{Marginal replay outperforms conditional replay when many samples can be generated}
From Fig.~\ref{fig:all_task_accuracy}, it can be observed that
marginal replay outperforms conditional replay by a small margin. This comes at the price of having to generate a large number of samples, which will become unfeasible if many classes are involved in the retraining.\\

\par\noindent
\textbf{Conditional replay is superior when few samples are generated} 
The results of Fig.~\ref{fig:bal} show that conditional replay is superior to marginal replay when generating fewer samples at each sub-task (more precisely: $0.1tN$ samples instead of $tN$, for sub-task $t$ and number of new samples per sub-task N). This can be understood quite easily: since we generate only $0.1tN$ samples instead of $tN$ samples at each sub-task, marginal replay produces an unbalanced class distribution, see Fig.~\ref{fig:distribution}, which strongly impairs classification performance. This is a principal advantage that conditional replay has over marginal replay: generating balanced class distributions while having much more control over the number of generated samples.
\\

\section{Conclusions}
\textbf{Summary} We have proposed several of performing continual learning with replay-based models and empirically demonstrated (on novel benchmarks) their merit w.r.t. the state of the art, represented by the EWC method. A principal conclusion of this article is that conditional replay methods show strong promise because they have competitive performance, 
and they impose less restrictions on their use in applications. Most notably, they
can be used at constant time complexity, meaning that the number of generated samples 
does not need to increase over time, which would be problematic in applications with many sub-tasks and real-time constraints. \\

\par\noindent
\textbf{Concerning the benchmarks} While one might argue that MNIST and FashionMNIST are too simple for a meaningful evaluation, this holds only for non-continual learning scenarios. In fact, recent articles \cite{pfuelb2019a} show that MNIST-related CLTs are still a major obstacle for most current approaches to continual learning under realistic conditions. So, while we agree that MNIST and FashionMNIST are not suitable benchmarks in general anymore, we must stress the difficulty of MNIST-related CLTs in continual learning, thus making these benchmarks very suitable indeed in this particular context. The use of intrinsically more complex benchmarks, such as CIFAR,SVHN or ImageNet is at present not really possible since generative methods are not really good enough for replaying these data \cite{2018arXiv181209111L}.

\par\noindent
\textbf{Next steps} Future work will include a closer study of conditional replay: in particular, we would like to better understand why they exhibit better performance w.r.t marginal replay in cases
where the number of generated samples is restricted to be low. In addition, it would be interesting to study the continual learning behavior of conditional replay models when a fixed number of generated samples is imposed at each sub-task, for various CLTs. The latter topic is interesting because the success of replay-based continual learning methods in applications will depend on whether the number of generated samples (and thereby time and memory complexity) can be reduced to manageable levels.\\

\par\noindent
\textbf{Observations} An interesting point is that the disjoint type CLTs pose enormous problems to conventional machine learning architectures, and therefore represent a very useful benchmark for continual learning algorithms. If each of a CLT's sub-tasks contains a single visual class, training them one after the other will induce no between-class discrimination at all since 
every training step just "sees" a single class. Replay-based methods nicely bridge this
gap, allowing continual learning while allowing between-class discrimination. To our mind, any application-relevant algorithm for continual learning therefore must include some form of experience replay.\\

\par\noindent
\textbf{Outlook} Ultimately, the goal of our research is to come up with replay-based models where the 
effort spent on replaying past knowledge is small compared to the effort of training with new samples, which will require machine learning models that are, intrinsically, less prone to catastrophic forgetting than DNNs are.

\bibliographystyle{splncs04}
\bibliography{samples.bib,lll}

\label{ap:first}

\end{document}

%% file: Tables/hyperparameters.tex
\begin{table*}[ht!]
\centering

  \caption{Hyperparameters for MNIST and Fashion MNIST all models (all CL settings have the same training hyper parameters with Adam)}
  \label{tab:hyperparams}
  \begin{tabular}{ccccccc}
    \hline 
    Method &
    Epochs&
    LR Classifier&
    LR Generator&
    beta1&
    beta2&
    Batch Size\\
    \hline
    
    Marginal Replay     & 
    25 & 
    0.01& 
    2e-4 & 
    5e-1& 
    0.999& 
    64 \\ 
    \hline
    
    Conditional Replay     & 
    25 & 
    0.01& 
    2e-4 & 
    5e-1& 
    0.999& 
    64 \\ 
    \hline
    
    EWC     & 
    25 & 
    0.01& 
    -& 
    5e-1& 
    0.999& 
    64\\ 
    \hline
    
    

    
    
    \hline
\end{tabular}

\end{table*}